\documentclass[letterpaper,twocolumn]{article} 
\usepackage{times}  % DO NOT CHANGE THIS
\usepackage{helvet}  % DO NOT CHANGE THIS
\usepackage{courier}  % DO NOT CHANGE THIS
\usepackage[hyphens]{url}  % DO NOT CHANGE THIS
\usepackage{graphicx} % DO NOT CHANGE THIS
\usepackage{caption} % DO NOT CHANGE THIS AND DO NOT ADD ANY OPTIONS TO IT
\usepackage{algorithm}
\usepackage{algorithmic}
\usepackage{amsfonts,amssymb} 
\usepackage{color}
\usepackage{multicol,multirow}
\usepackage{subcaption}
\usepackage{graphicx}
\usepackage{amsmath}

\newcommand{\etal}{\textit{et al. }}

\usepackage{newfloat}
\usepackage{listings}
\DeclareCaptionStyle{ruled}{labelfont=normalfont,labelsep=colon,strut=off} % DO NOT CHANGE THIS
\floatstyle{ruled}
\newfloat{listing}{tb}{lst}{}
\floatname{listing}{Listing}
\title{Realistic Speech-to-Face Generation with Speech-Conditioned \\ Latent Diffusion Model with Face Prior}

\author{Jinting Wang\textsuperscript{1},\   
Li Liu\textsuperscript{1}\thanks{Corresponding Author: avrillliu@hkust-gz.edu.cn.}\ ,\  
Jun Wang\textsuperscript{2},\  
Hei Victor Cheng\textsuperscript{3}
\\
\textsuperscript{1}{The Hong Kong University of Science and Technology (Guangzhou)}\\
\textsuperscript{2}{Tencent AI Lab}\\
\textsuperscript{3}{Aarhus University}\\
}
\begin{document}

\maketitle

\begin{abstract}

 %existing methods often suffer from issues such as %\textcolor{red}{unnatural distortions, and poor image quality}. For example, SOTA methods employing GAN-based architecture lack stability and cannot generate realistic images due to ...
 Speech-to-face generation is an intriguing area of research that focuses on generating realistic facial images based on a speaker’s audio speech. However, state-of-the-art methods employing GAN-based architectures lack stability and cannot generate realistic face images.
To fill this gap, we propose a novel speech-to-face generation framework, which leverages a \textbf{S}peech-\textbf{C}onditioned \textbf{L}atent \textbf{D}iffusion \textbf{M}odel, called \textbf{SCLDM}. To the best of our knowledge, this is the first work to harness the exceptional modeling capabilities of diffusion models for speech-to-face generation.
Preserving the shared identity information between speech and face is crucial in generating realistic results. Therefore, we employ contrastive pre-training for both the speech encoder and the face encoder. This pre-training strategy facilitates effective alignment between the attributes of speech, such as age and gender, and the corresponding facial characteristics in the face images.
Furthermore, we tackle the challenge posed by excessive diversity in the synthesis process caused by the diffusion model. To overcome this challenge, we introduce the concept of residuals by integrating a statistical face prior to the diffusion process. This addition helps to eliminate the shared component across the faces and enhances the subtle variations captured by the speech condition.
Extensive quantitative, qualitative, and user study experiments demonstrate that our method can produce more realistic face images while preserving the identity of the speaker better than state-of-the-art methods.
Highlighting the notable enhancements, our method demonstrates significant gains in all metrics on the AVSpeech dataset and Voxceleb dataset, particularly noteworthy are the improvements of 32.17 and 32.72 on the cosine distance metric for the two datasets, respectively.

%One of the main challenges in this task is to 
%This paper presents a cutting-edge approach for automatic speech-to-face conversion using a diffusion model based on face priors and pretrained speech features. By leveraging a state-of-the-art pretrained speech model, our method captures the intricate relationship between speech signals and facial expressions, enabling the synthesis of realistic and expressive facial animations. Incorporating face priors provides prior knowledge about facial structures, dynamics, and appearance, resulting in enhanced realism and fidelity in the generated animations. The diffusion model acts as a powerful framework for propagating information from the pretrained speech features to generate coherent and temporally consistent facial movements. Through extensive experiments on diverse datasets and a user study, we demonstrate that our approach outperforms existing methods in terms of visual quality, temporal coherence, and perceptual realism. This work showcases significant advancements in automatic speech-to-face conversion and holds promising implications for applications in virtual avatars, multimedia communication, and the entertainment industry, offering users an immersive and interactive experience.
\end{abstract}

 %\begin{table*}[t]
 %    \centering
 %    \begin{tabular}{l|cccc}
 %    \hline
  %        & Wav2Pix  & Speech2Face& Wen $\etal$ & Ours \\
 %        \hline
  %        Audio type&  Waveform& Spectrogram& Mel-Spectrogram& Spectrogram\\
  %        Auxiliary information& No& No& Identity& No\\
 %         Framework& GAN& CNN& GAN& LDM\\
 %         Speech-face alignment& No& No&No& Yes\\
  %        Face prior &No& No& No&Yes\\
  %        \hline
  %   \end{tabular}
  %   \caption{Comparison of details with the SOTA speech-to-face generation methods}
  %   \label{tab:comparison details with others}
 %\end{table*}

 \begin{figure*}[t]
\centering
\includegraphics[width=1\textwidth]{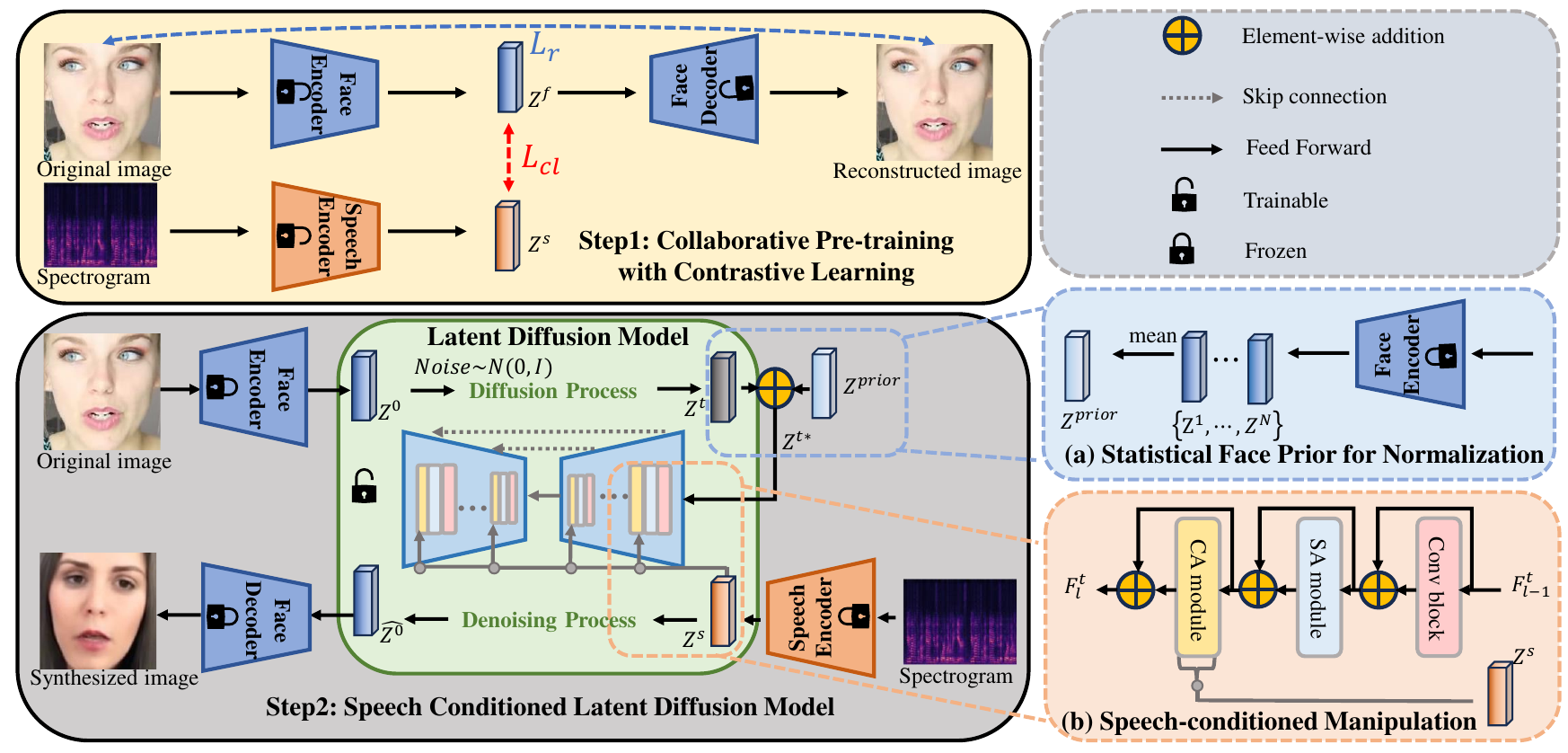} % Reduce the figure size so that it is slightly narrower than the column.
\caption{Details of the diffusion-based Speech-to-Face generation pipeline. It consists of two stages: \textbf{(1) Collaborative Pre-training with Contrastive Learning}. The speech encoder and face encoder take the audio and face image as input, respectively, to learn the representations, the captured embedding of face and speech are aligned by contrastive learning. \textbf{(2) SCLDM.} The aligned face embedding is gradually destroyed by Gaussian noise in the diffusion process within iterative time steps. Then it is combined with common face stuff provided by face prior and fed into LDM for face generation. The aligned speech embedding is utilized as conditioned variation to manipulate the face generation in the training process of LDM. }
\label{fig2}
\end{figure*}
 
\section{Introduction}

Generating realistic face portraits of speakers based on their audio speech has numerous applications, such as virtual anchors, teleconferencing, surveillance, and digital human animation. The field of speech-to-face generation aims to establish a meaningful mapping between audio and visual faces, resulting in high-quality and genuine face images. Previous studies have explored the relationship between human voice and facial structures, supporting the feasibility of speech-to-face generation \cite{belin2004thinking,kamachi2003putting,wang2020attention}.  Factors such as facial bone structure, joint configuration, and the tissues involved in sound production are closely intertwined with the shape and size of facial organs. Additionally, various factors including genetics, biology, and the environment impact both voice and face characteristics. For instance, gender, age, and ethnicity have been found to significantly influence both speech and facial appearance \cite{kamachi2003putting, wang2022acoustic, kwasny2021gender,wang2022residual}. 

Despite the potential of leveraging shared characteristics between face and voice for a speech-to-face generation, it remains a complex task due to the diverse nature of human faces and the various speaking styles they encompass.
There are two primary challenges: 1) How can a model learn the common characteristics shared by voice and face? 2) How can a model produce visually authentic facial images while preserving attribute details?
To address the first challenge, some research utilizes extracted speech features to generate face images based on various existing generative models
\cite{oh2019speech2face,wen2019face,duarte2019wav2pix,fang2022facial}. 
Although delicate designs have been introduced to align the speech features and face features, a substantial gap persists between speech embedding and visual embedding, resulting in subpar face generation. Furthermore, to achieve speech-to-face conversion, GAN-based architectures are mainly employed in previous studies. However, methods based on GANs require simultaneous optimization of a generator and a discriminator, such a training process lacks stability and is prone to an effect known as mode collapse \cite{dhariwal2021diffusion}. Therefore, the generated speaker faces are of limited image quality.

Inspired by the Stable Diffusion \cite{rombach2022high}, which employs latent diffusion models (LDMs) to achieve high-quality image generation, we propose a novel speech-to-face generation framework, which incorporates a \textbf{S}peech-\textbf{C}onditioned \textbf{LDM} (\textbf{SCLDM}). Our framework is designed to address the limitations observed in previous works and to advance the progress of speech-to-face generation. Moreover, we aim to achieve state-of-the-art (SOTA) face generation quality that aligns seamlessly with the attributes of the spoken speech. To achieve this, we explore speech-conditional face manipulations with LDMs, which have not been explored before. Specifically, an LDM conditioned on the latent speech embedding is learned for face image generation. For accurate speech-conditioned face manipulations, we leverage the face-aligned speech latent embedding space pre-trained by contrastive learning as the condition.
Moreover, we introduce a statistical face prior to the LDM. This statistical prior serves as the shared component of the faces and assists the LDM to emphasize the subtle changes present in speech while alleviating its diversity for identity attribute preservation.

The main contributions of this work are as follows:
\begin{itemize}
    \item A novel speech-conditioned diffusion-based network is proposed for speech-to-face generation. To the best of our knowledge, this is the first attempt to develop an LDM for this task. 
    \item To achieve accurate and delicate speech-conditioned face manipulation, a contrastive pre-training is employed for aligning speech and face representations. Then the aligned speech latent embedding is utilized as a condition to generate a face image corresponding to the speech.
    \item To enhance the speech manipulation, a statistical face prior is introduced to the diffusion model, which allows for generating a more realistic face image conforming to the corresponding speech.
    \item Extensive experiments are conducted to validate the performance of our proposed method, the results show that SCLDM achieves the best among all speech-to-face generation methods. 
\end{itemize}

\section{Related Work}

\subsection{Diffusion Model}
Diffusion models have demonstrated their SOTA generation performance in various tasks, including image generation \cite{rombach2022high,dhariwal2021diffusion}, speech generation \cite{kim2022guided,lee2023imaginary}, and video generation \cite{luo2023videofusion,blattmann2023align}. In diffusion-based frameworks, one of the major concerns is to improve the efficiency of diffusion models due to their iterative generation process in a high-dimensional data space. The latent diffusion model is one of the solutions that apply diffusion models in a small latent space \cite{rombach2022high}, which was first presented in image generation \cite{rombach2022high}. An LDM model is trained as a flexible image generator with general conditioning inputs such as semantic map, text, and image. This approach inspired further study on conditional LDMs in different domains, such as text-to-audio generation \cite{huang2022prodiff}, text-to-image generation and editing \cite{saharia2022photorealistic}, text-to-video generation and editing \cite{blattmann2023align}, conditional image-to-video generation \cite{ni2023conditional}, and audio-to-video generation \cite{zhu2023taming}.
In this work, our focus is on leveraging the LDM models and proposing a diffusion-based framework for speech-to-face generation.

\begin{figure*}[t]
\centering
\includegraphics[width=0.8\textwidth]{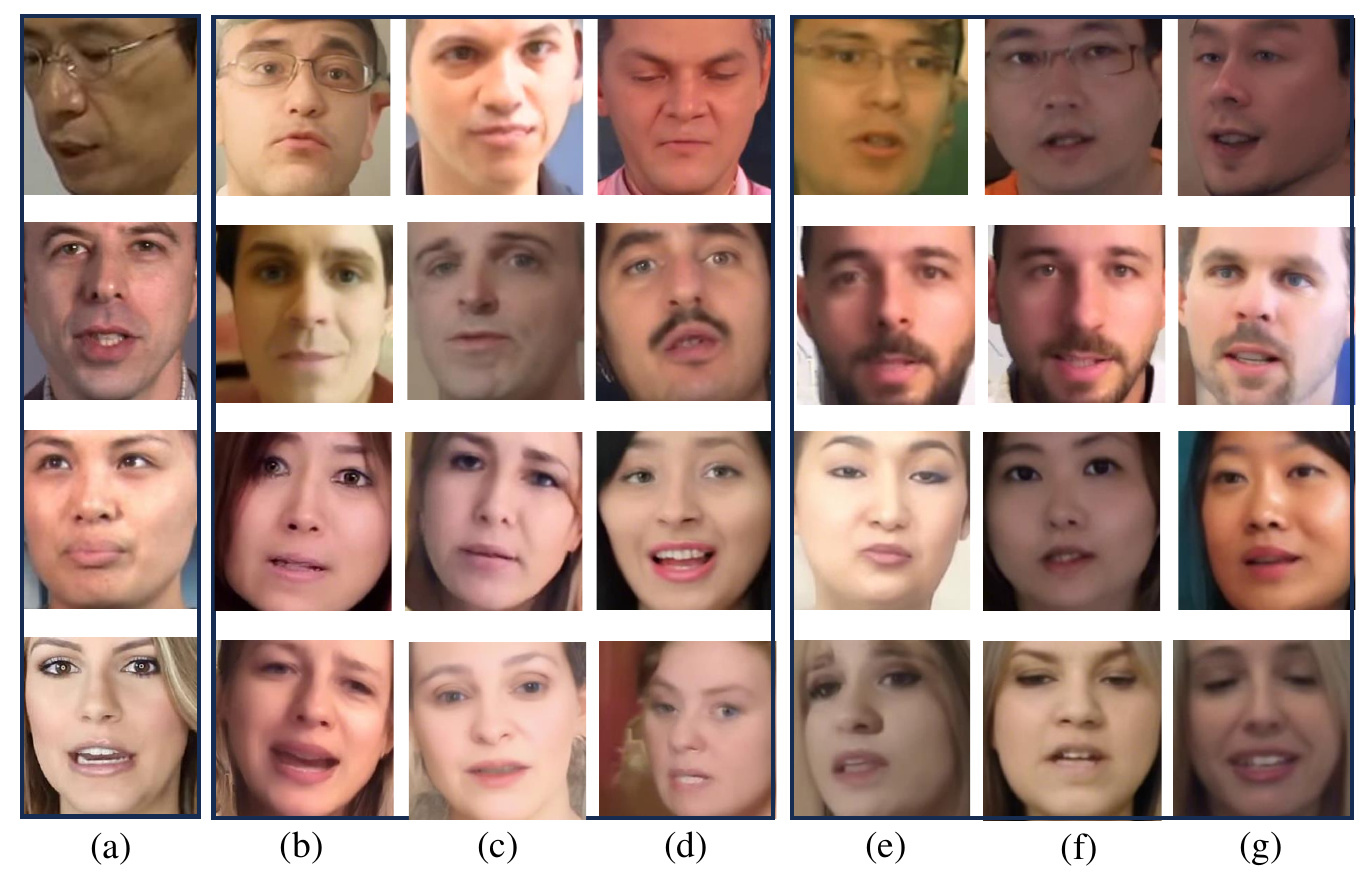} % Reduce the figure size so that it is slightly narrower than the column.
\caption{Qualitative comparison of the face prior. (a) Face images cropped from the video frame. (b-c) Top-3 generated faces of the same speech without prior normalization. (e-g) Top-3 generated faces of the same speech with prior normalization. }
\label{fig:ablation_prior}
\end{figure*}

\subsection{ Speech-to-Face Generation}
Audio-visual cross-modal learning, particularly speech-to-face generation, has gained significant attention in recent years. The goal of speech-to-face generation is to generate realistic facial appearances that correspond to the audio speech input. Many existing methods in this field employ GAN-based frameworks.
one such framework is Wav2Pix \cite{duarte2019wav2pix}, which proposes a speech-conditioned face generation framework. It consists of a speech encoder, a generator, and a discriminator. Wav2Pix is relatively simplistic and overlooks the preservation of identity information during the generation process.
To address the preservation of identity information, Wen et al. \cite{wen2019face} design a network capable of generating faces from voices by matching the identities of the generated faces with those of the speakers. This approach aims to preserve certain biometric characteristics of the speakers. Similarly, Fang et al. \cite{fang2022facial} incorporate speaker identity information to learn identity representations across different modalities.
While explicitly modeling the identity relevance between audio and visual modalities is beneficial for ensuring the authenticity of generated face images, it has limitations when attempting to generate faces of different identities. On the other hand, Choi et al. \cite{choi2019inference} propose a two-stage framework comprising an inference stage for cross-modal matching and a generation stage for speech-conditioned face generation. This approach allows for more flexibility in generating faces with different identities.
In the work of Oh et al. \cite{oh2019speech2face}, a one-stage method is proposed, incorporating a multivariate mixing loss function to enforce consistency between speech and face features. This facilitates the learning of shared attribute information between the voice and the face.
The existing methods for speech-to-face generation have shown some progress, but they still have limitations, particularly in terms of generation quality. Additionally, the alignment between speech and face modalities has not been adequately explored, which can impact the overall coherence and accuracy of the generated results.
In this work, we propose a speech-to-face generation network that employs LDM as the generation model, and speech-face alignment is conducted to enhance modality matching.

\section{Proposed Method}

In this work, we propose a diffusion-based framework for generating realistic face images from speech input, which exploits the SOTA LDM for speech-conditioned face synthesis. Considering shared information preservation between speech and face image modality is crucial for realistic image generation, we employ a pre-training step that utilizes a contrastive learning strategy to facilitate alignment between speech and face embedding.
The aligned speech embedding is then used as a conditioning input to manipulate the face generation process of the LDM. 
Furthermore, we introduce a statistical face prior into the latent code of the LDM. This face prior provides a face template that guides the model’s attention towards variations corresponding to the input speech. 

Figure \ref{fig2} provides detailed insights of the proposed diffusion-based speech-to-face generation framework, SCLDM, it can be seen that it consists of two stages. In the first stage, the face encoder, face decoder, and speech encoder are collaboratively pre-trained using a contrastive learning approach. This pre-training step ensures that the model learns to align speech and face embeddings effectively. In the second stage, the pre-trained face encoder and speech encoder are used to encode aligned embeddings for training the speech-conditioned LDM. The pre-trained face decoder is kept for the testing stage, where it is used to reconstruct the face images.

\subsection{Collaborative Pre-training with Contrastive Learning}
\label{subsec:Contrastive Speech-Face pre-training }
Inspired by the great success of contrastive pre-training in various cross-modal applications \cite{afham2022crosspoint,parelli2023clip}, contrastive pre-training is employed in this stage to facilitate speech-face alignment.

As shown in Figure \ref{fig2}, given an audio clip of a speaker and the corresponding face image, a speech encoder $E_{Speech}(.)$ and a face encoder $E_{Face}(.)$ are employed to extract the speech embedding $Z^s \in \mathcal{R}^d$ and face embedding $Z^f \in \mathcal{R}^d$, respectively, where $d$ is the dimension of the embedding vectors.
To align the speech embedding and face embedding, a symmetric cross-entropy loss \cite{radford2021learning} is applied, leveraging contrastive learning techniques. Specifically, we use VGGFace \cite{qawaqneh2017deep} as the face encoder, and a model combined with CNN (the speech encoder architecture in Speech2Face \cite{oh2019speech2face}) and CBAM module \cite{woo2018cbam} as the speech encoder. Since we apply the diffusion model in a latent space, a face decoder is required to reconstruct face images from the latent representation. A CNN-based model symmetrical to the VGGFace is designed as the face decoder $D_{Face}(.)$. 
Finally, a combination of MAE loss and LPIPS loss \cite{zhang2018unreasonable} is used as the reconstruction loss. 
%\hvc What are MAE and LPIPS loss?\fin

The objective function of the collaborative pre-training $L_C$ is defined as:
\begin{equation}
    L_C = L_{cl} + L_r,
\end{equation}
where $L_{cl}$ and $L_r$ denote the contrastive loss and reconstruction loss, respectively. 
%The training details are provided in \textcolor{red}{ Appendix.}
After training, the speech representation $Z^s$ of a random audio sample is used for providing shared-attributes information.

\subsection{Explicit Face Prior as Normalization }
\label{subsec: face priors}

\textbf{Motivations.}
While the speech-conditioned LDM can generate positive results, we observe that it occasionally falls short in producing realistic face images that accurately match the speaker attributes. 
This can be seen in Figure \ref{fig:ablation_prior}, 
where an LDM trained with speech condition is employed to create speaker portraits. However, the outputs originating from the same speech clip exhibit diverse characteristics, rendering them distinct from one another. This diversity has a detrimental impact on the visual quality, hindering their use in real-world applications. 
This shortcoming is attributed to the inherent diversity present in the  latent space learned by the LDM, which is a result of the complex data distribution. During the denoising process for face synthesis, the latent is randomly sampled from the latent space, resulting in the variance observed in the generated images. This observation suggests that relying solely on speech conditions to exert control is insufficient for generating the desired face images that accurately correspond to the speech clips. This highlights the necessity for supplementary mechanisms to guide the generation process.

\textbf{Latent Diffusion with Face Prior as Normalization.} 
%There is a huge workload to make an effort to train data since a large amount of face-speech pairs are collected from the wild. 
%Considering the complex data distributed in the latent space of LDM, and the latent code $Z^t$ used to generate a face image in an iterative denoising process are sampled from it, 
To address the aforementioned issue and generate more realistic face images conforming to the speech, we propose to explicitly introduce a statistical face prior to the LDM for normalization purposes. This approach exploits the concept of residuals to eliminate the shared component found in the faces and effectively emphasizes the subtle variations that are captured by the speech embedding.

As illustrated in Figure \ref{fig2}, the statistical face prior $Z^{prior}$ is combined with the latent code as a weighted sum:
\begin{equation}
    Z^{t*} = Z^t + \beta Z^{prior}.
\end{equation}
%Then $Z^{t*}$ serves as a template in the iterative denoising process of LDM.  
Incorporating the face prior maintains the fundamental mechanisms of the LDM. The inclusion of the prior enriches the LDM's learning process by guiding it to learn the difference between the input face images and the introduced face prior within the diffusion process. In the denoising process, we effectively alter the distribution of the latent code that we sample from, shifting from $\mathcal{N}(\textbf{0},\textbf{I})$ to $\mathcal{N}(\beta Z^{prior},\textbf{I})$. The choice of weight $\beta = 0.01$ is based on empirical knowledge, and we will provide an ablation study of the weight $\beta$ in the experiment part.
%against interfering factors and normalize the generated results.

The face prior is constructed by averaging the features extracted from the pre-trained face encoder $E_{Face}$ on given gender-balance data:
\begin{equation}
    Z^{prior} = \frac{1}{N} \sum_{1}^{N} E_{Face}(f),
\end{equation}
where $f$ denotes a face image and $N$ is the total number of face images. 
Through experiments we observe that when $N$ gradually increases, the face prior tends to converge, suggesting that the calculated prior becomes representative of the shared characteristics. As indicated in Table \ref{tab:face prior num}, we take $N=10000$ in this work.

\begin{table}[h]
    \centering
    \begin{tabular}{ccc}
    \hline
      $N_1$ & $N_2$ & $L1(N_1, N_2)$\\
      \hline
        100 &500 & 11.86\\
        500& 1000& 5.21\\
        1000& 5000&3.35 \\
        5000& 10000&1.28 \\
        10000& 15000& 0.67\\
        \hline
    \end{tabular}
    \caption{$L1$ distance between face prior calculated with different number $N$ of face image.}
    \label{tab:face prior num}
\end{table}

With the incorporation of the face prior into LDM as normalization, SCLDM demonstrates improved precision. It is capable of synthesizing face images that better preserve the identity of the speaker, as depicted in Figure \ref{fig:ablation_prior}.

\subsection{Speech-Conditioned Latent Diffusion Model with Face Prior}
\label{subsec:Speech-conditioned Latent Diffusion Model}

In speech-to-face generation, the speaker portrait is synthesized given the speech clip as the condition. With speech-conditioned LDM, we are interested in generating face images conforming to the speech. 
%under the manipulation of conditioning vector $Z^s$, which is extracted by the pre-trained speech encoder $E_{Face}$. 
Specifically, in the training stage, the face image is embedded into latent representation $Z^f$ ($Z^0$ in diffusion process) by the trained face encoder $E_{Face}$, and then the face embedding is destroyed into a noised vector $Z^t$ with Gaussian distribution in the diffusion process after $t$ time steps, which is denoted as:
\begin{equation}
    Z^t:=\alpha^t*Z^0+(1-\alpha^t)*\epsilon,
\end{equation}
where $\epsilon \sim \mathcal{N}(\textbf{0},\textbf{I})$ denotes the injected noise, and $\alpha^t$ represents the noise level at the $t$ time step. 
With the prior normalization mentioned above, $Z^t$ would be transformed into $Z^{t*}$ by introducing the face prior $Z^{prior}$.
For noise estimation, a denoising model compromised of the UNet backbone and attention mechanism is employed in the denoising process. The condition vector $Z^s$ captured by the pre-trained speech encoder $E_{Speech}$ would be mapped to the intermediate layers of the UNet via a cross-attention mechanism. As shown in Figure \ref{fig2}, for a specific layer $l$, the speech-conditioned manipulation is that the layer input $F^t_{l-1}$ successively processed by a convolution block and a self-attention module, and then interacted with the condition vector $Z^s$ via a cross-attention module. The speech-conditioned operation is defined as:
\begin{equation}
    \mathrm{CA}(Q, K, V) = \mathrm{Softmax}\left(\frac{QK^T}{\sqrt{d}}\right)V,
\end{equation}
where $\mathrm{CA}$ is the abbreviation of cross attention module, $Q$ is the linear transformation of the output of self-attention module (SA). While $K$ and $V$ are transformed from condition vector $Z^s$ by two linear layers.
 Based on speech-face pairs, the conditional LDM is trained via 
\begin{equation}
    L_{LDM} := \mathbb{E}_{\epsilon, Z^{t*}, t} \left[\left\|\epsilon-\epsilon_\theta\left(Z^s, Z^{t*}, t\right)\right\|^2\right],
\end{equation}
where $\epsilon_\theta$ denotes the optimized denoising model, and $\theta$ denotes its parameters.

In the generation process, starting from injecting face prior  $Z^{prior}$ into Gaussian noise $Z^t \sim \mathcal{N}(\textbf{0},\textbf{I})$, the trained denoising model gradually generates latent face $\hat{Z^0}$ with the speech condition, and then it is fed into the trained face decoder and the face reconstruction is obtained. Details of the training and generation process are provided in the supplementary materials.
%\hvc
%Details of the training and generation process are provided in Appendix.
%\f
\begin{table*}[t]
 \centering
    \begin{tabular}{l|ccc|cc}
    \hline
    
        \multirow{2}{*}{Method} & \multicolumn{3}{c}{Feature Similarity} \vline & \multicolumn{2}{c}{Identity Preservation} \\
    \cline{2-6}
          & L1 $\downarrow$ & L2 $\downarrow$ & cos $\downarrow$ &
          gender $\left (\%\right) \uparrow$ & age $\left (\%\right) \uparrow$ \\
    \hline
    Wav2Pix&144.72& 24.32&82.51& 67.4&41.3   \\
    Speech2Face& 67.18&3.94 &46.97 &95.6 &65.2  \\
    SCLDM (Ours)& 35.01&1.48 &12.81 &98.8 &84.5  \\
    \hline
    \end{tabular}
    \caption{Comparison results on AVSeech dataset. }
    \label{tab:comparison resuls on AVSpeech}
\end{table*}

\begin{table*}[t]
    \centering
   
    \begin{tabular}{l|ccc|cc}
    \hline
    
        \multirow{2}{*}{Method} & \multicolumn{3}{c}{Feature Similarity} \vline & \multicolumn{2}{c}{Identity Preservation} \\
    \cline{2-6}
          & L1 $\downarrow$ & L2 $\downarrow$ & cos $\downarrow$ &
          gender $\left (\%\right) \uparrow$ & age $\left (\%\right) \uparrow$ \\
    \hline
    Wav2Pix&137.58 &22.19 &79.36 & 74.5&49.6   \\
    Speech2Face& 66.46&2.77 &44.38 &96.1 & 69.4 \\
    Wen $\etal$& 59.82& 2.41& 42.54&97.4 & 72.5 \\
    SCLDM (Ours)&27.10 &1.09 &11.54 &99.4 &88.6  \\
    \hline
    \end{tabular}
    \caption{Comparison results on VoxCeleb dataset. }
   
    \label{tab:comparison results on VoxCeleb}
\end{table*}

\begin{figure}[t]
\centering
\includegraphics[width=0.95\columnwidth]{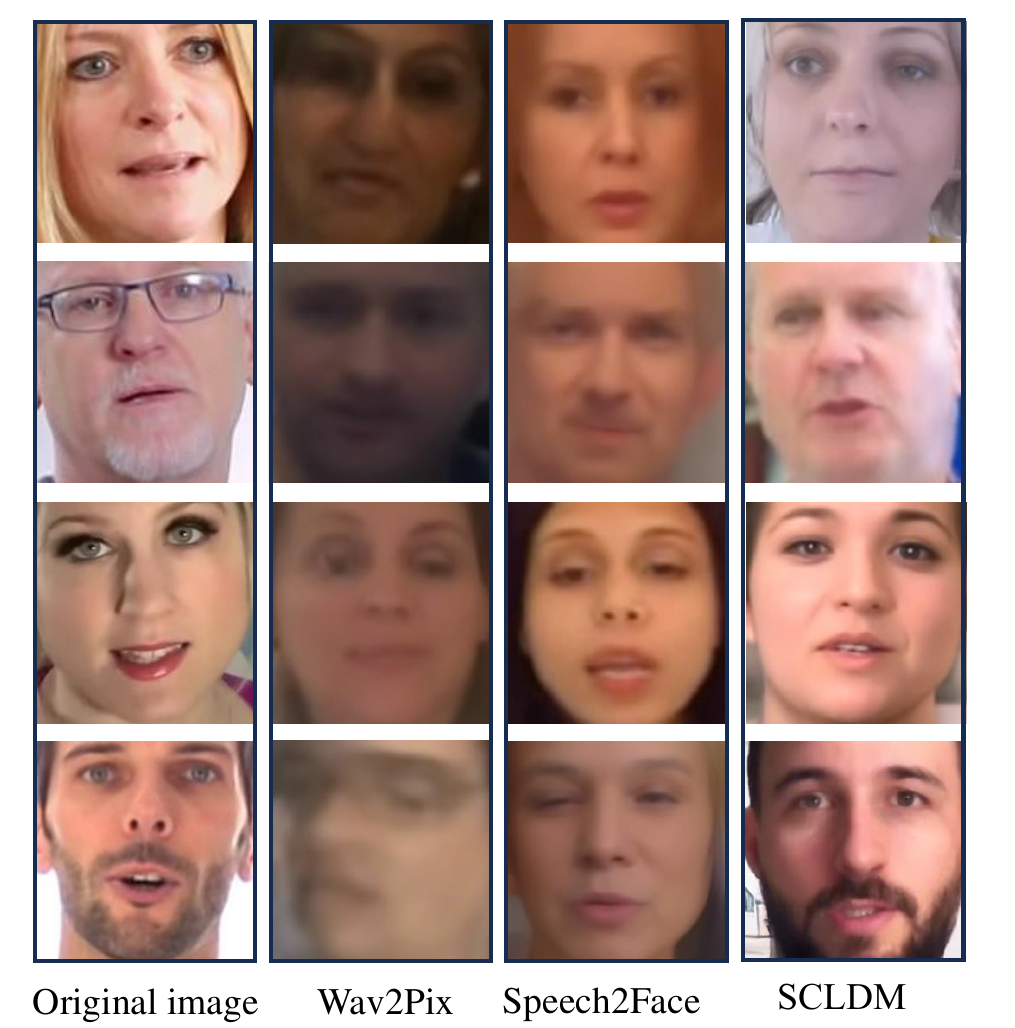} % Reduce the figure size so that it is slightly narrower than the column.
\caption{Qualitative comparison of our model and SOTA methods on the AVSpeech dataset. }
\label{fig:comparative_Avspeech}
\end{figure}

\begin{table*}[t]
    \centering
   
    \begin{tabular}{ccc|ccc|cc}
    \hline
    
        \multicolumn{3}{c}{Method} & \multicolumn{3}{c}{Feature Similarity} \vline & \multicolumn{2}{c}{Identity Preservation} \\
    \hline
         base & CP & PN & L1 $\downarrow$ & L2 $\downarrow$ & cos $\downarrow$ &
          gender $\left (\%\right) \uparrow$ & age $\left (\%\right) \uparrow$ \\
    \hline
     $\checkmark$& & &56.39 &3.30&29.83&95.9 & 74.7 \\
    $\checkmark$&\checkmark& &47.61 &2.69 &21.79&97.2 & 82.6 \\
    $\checkmark$& & \checkmark& 44.27& 2.38&20.41 &96.4&80.3 \\
    $\checkmark$&$\checkmark$&$\checkmark$& 35.01&1.48 &12.81 &98.8 &84.5  \\
    \hline
    \end{tabular}
    \caption{Ablation results on AVSpeech dataset. Abbreviations ``CP" and ``PN" denote Collaborative Pre-training and Prior Normalization, respectively.}
   
    \label{tab:ablation results on VoxCeleb}
\end{table*}

\begin{figure}[h]
    \centering
    \includegraphics[width=0.9\columnwidth]{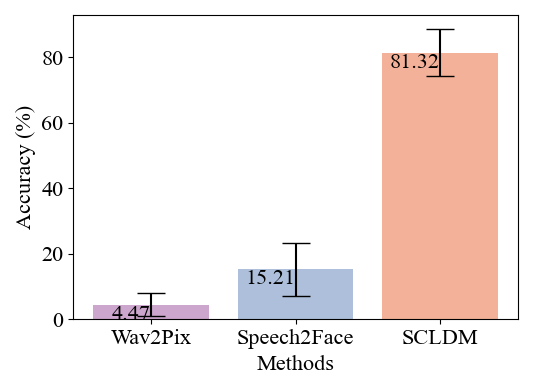}
    \caption{Results of the user study. Among the three methods, our method achieves the highest accuracy for users evaluation, in terms of image quality and identity preservation.}
    \label{fig:user_study}
\end{figure}

\begin{figure}[htbp]
\centering
\includegraphics[width=\columnwidth]{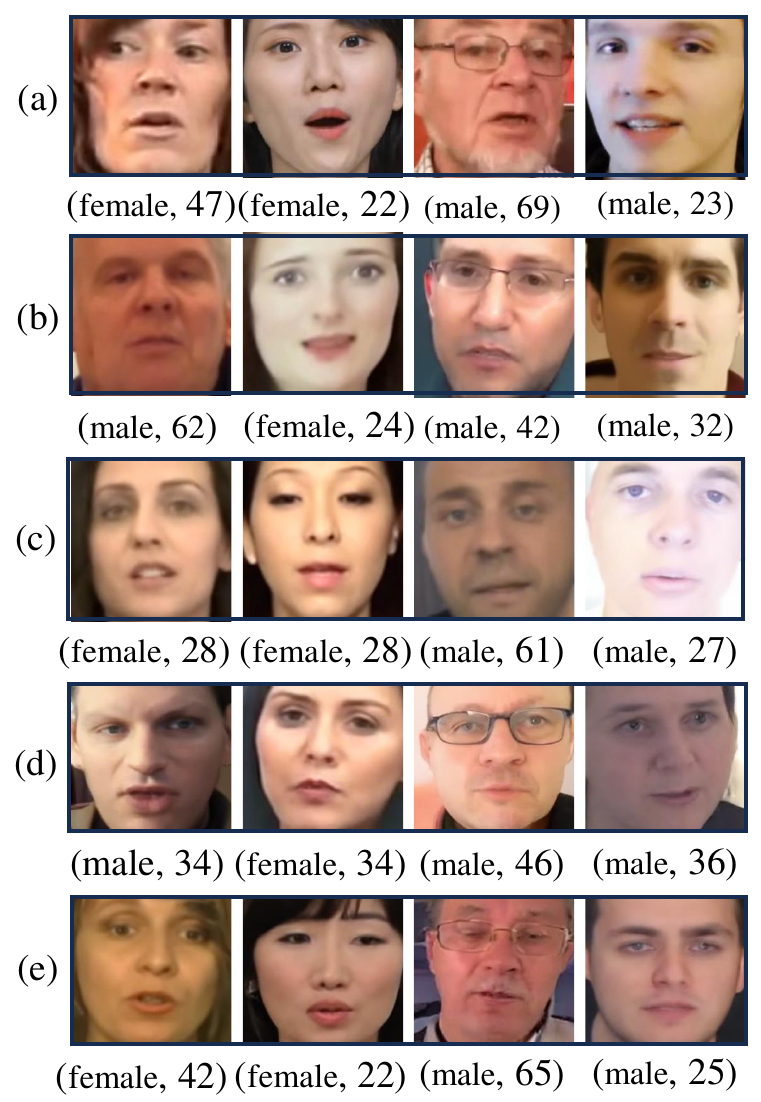} % Reduce the figure size so that it is slightly narrower than the column.
\caption{Qualitative comparison of ablation studies. (a) Original face images cropped from the video frame. (b) Generated face by speech-conditioned LDM (base). (c) Generated face by speech-conditioned LDM with collaborative pre-training. (d) Generated face by speech-conditioned LDM with face prior normalization. (e) Generated face by conditional LDM with collaborative pre-training and face prior normalization (SCLDM).}
\label{fig:ablation_cl}
\end{figure}

\section{Experiments}
\subsection{Dataset}
Throughout the experiments, two datasets are used.
\textbf{AVSpeech.} AVSpeech \cite{ephrat2018looking} is a large-scale audio-visual dataset collected from YouTube, which consists of 2.8m video clips. It is the significant diversity present in the included face images which were extracted from videos captured “in the wild”.

\noindent \textbf{VoxCeleb.} VoxCeleb \cite{nagrani2017voxceleb} contains 1251 speakers, which span a wide range of ethnicities, accents, professions, and ages. The nationality and gender of each speaker are also provided in the dataset. 

\subsection{Evaluation Metrics}

\textbf{Feature Similarity.} Following \cite{oh2019speech2face}, we measure the cosine, L1, and L2 distance between the features of the true face image and generated face image extracted by VGGFace \cite{qawaqneh2017deep}, a trained face recognition network.

\noindent\textbf{Identity Preservation.} We employ Face++\footnote{https://www.faceplusplus.com/attributes.}, a commercial API for face attribute recognition, to evaluate face attributes, including age and gender. Note that the age accuracy is computed within a 10-years range.

% \noindent\textbf{Speech-Face Matching Score} We compute the cosine similarity between the speech and generated face embeddings. A higher score indicates higher consistency between the generated face image and speech.

\subsection{Implementation Details}
Following Speech2Face \cite{oh2019speech2face}, we use 6 seconds of audio taken from the video clip and then process it into a spectrogram by STFT. The face images are resized to $256 \times 256$ pixels.
For collaborative pre-training, we set a learning rate for the face encoder and face decoder of 0.0001, and 0.001 for the speech encoder. 
We adopt LDM \cite{rombach2022high} as a generator since it achieves a good balance between quality and speed. In the LDM training stage, the face encoder and speech encoder are frozen, and optimization is performed with a learning rate of 2e-5.

\subsection{Comparative Study}
 We compare our proposed method with three SOTA speech-to-face generation methods, i.e., Wav2Pix \cite{duarte2019wav2pix}, Speech2Face \cite{oh2019speech2face}, and the work of Wen $\etal$ \cite{wen2019face}.
 We conduct experiments using the default settings and official implementations for Wav2Pix \cite{duarte2019wav2pix} and the work of Wen $\etal$ \cite{wen2019face}. Since the code of Speech2Face \cite{oh2019speech2face} is not available, we reproduce it according to the paper.
 We only compare with Wen $\etal$ on the Voxceleb dataset since the identity information  of speakers is lacking in AVSpeech dataset.

 %We summarize the main differences between the compared methods and our approach in Table \ref{tab:comparison details with others}. 
 % We only compare with Wen $\etal$ on the Voxceleb dataset since the identity information  of speakers is lacking in AVSpeech dataset.

\noindent\textbf{Quantitative Comparison.} The comparison results on AVSpeech and VoxCeleb datasets are reported in Table \ref{tab:comparison resuls on AVSpeech} and Table \ref{tab:comparison results on VoxCeleb}, respectively. Our method outperforms all the competitors in all metrics. Specifically, the cosine distance of our method achieves 12.81 on AVSpeech test set and and 13.54 on VoxCeleb test set. The gender recognition accuracy achieves 98.8 and 98.4 on the two dataset. These results verify our effectiveness in speech-conditioned quality.

\noindent\textbf{Qualitative Comparison.} The qualitative comparison is presented in Figure \ref{fig:comparative_Avspeech}. It is observed that our framework is capable of synthesizing realistic outputs consistent with the attributes of speaker compared with Wav2Pix \cite{duarte2019wav2pix} and Speech2Face \cite{oh2019speech2face}.

\noindent\textbf{User Study.} We conduct a user study with 20 human evaluators to measure the effectiveness of the methods perceptually. We randomly sample 50 speech clips in the AVSpeech test set, then synthesize the speaker's face images given the speech. The evaluators are provided with the true face and the generated face images. They are asked to choose the best image based on 1) image realism, 2) identity preservation. Figure \ref{fig:user_study} showed the mean and standard deviation of the results, one can see that our framework significantly outperforms the existing STOA methods, which verifies the effectiveness in both image quality and identity preservation.

\subsection{Ablation Study}
\noindent\textbf{Ablation Experiment on Model Components.}
We conduct ablation studies on the AVSpeech dataset to validate the effectiveness of different components. 
The comparison results of different versions are listed in Table \ref{tab:ablation results on VoxCeleb}. It can be seen that accuracy gains a lot in both gender and age attributes with collaborative pre-training, which indicates that the identity information shared between face and speech is aligned and preserved by contrastive learning.  With face prior normalization, the feature distances between generated images and original faces are lower, which implies that the synthesized results present similar appearance as original faces. 

Additionally, we provide visual examples in Figure \ref{fig:ablation_cl} to illustrate the generated face images. It is evident that with the collaborative pre-training and prior normalization, the generated face images exhibit a similar appearance and attributes to the speaker in the corresponding speech.

\begin{figure}[htbp]
\centering
\includegraphics[width=0.8\columnwidth]{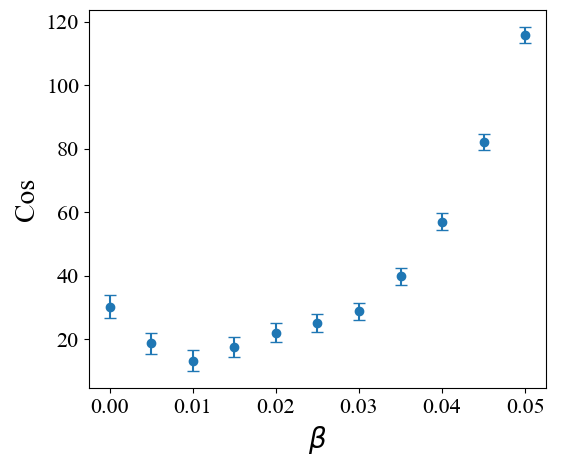} % Reduce the figure size so that it is slightly narrower than the column.
\caption{Ablation study on the weight $\beta$ of face prior normalization. }
\label{fig:ablation_prior2}
\end{figure}

\noindent \textbf{Ablation Experiment on Prior Normalization Weight.}
We conduct experiments to evaluate the impact of varying the weight of the face prior, $\beta$, during both the training and inference stages. The results of these experiments are shown in Figure \ref{fig:ablation_prior2}. In this analysis, we calculate the mean and variance of the cosine distance between the top-3 generated faces and the original images within the AVSpeech test set.
This finding suggests that incorporating the face prior with a weight of 0.01 during training and inference achieves the best quality for the generated faces.

\section{Conclusion and Discussion}
In this work, we propose a speech-conditioned latent diffusion model for speech-to-face generation, with the aim of generating realistic and identity-preserving face images given speech clips in real-world scenarios. 
With the assistance of contrastive pre-training, the speech encoder and face encoder provide aligned embedding that preserves the shared identity information, which enables effective manipulation of the generated faces.  
In addition, we utilize a statistical face prior in combination with a residual trick, which serves as normalization, removing the common components shared among faces. With the face prior, the LDM can better focus on capturing the subtle variations in the speech condition, resulting in more accurate and realistic face generation.
Extensive experiments demonstrate that our method achieves new SOTA performance on speech-to-face generation. We believe that our idea of utilizing latent diffusion model with a statistical face prior would be a good inspiration for future works in different face generation tasks.

\noindent \textbf{Limitations and Future Work.} 
Our framework focuses on exploiting diffusion models for speech-to-face generation, therefore, the face encoder and speech encoder are both built following the structures in previous work. However, the representation ability of the embeddings, particularly the speech embedding poses a high influence on conditioned generation. Future work should focus on finding better embeddings aligning speech and face features. 
A possible direction is to employ a large speech model to find audio signal embedding.
On the other end, the face prior introduced in our work is combined with the latent code of diffusion model with equal weights across samples, which may affect the diversity when the speech condition has limited variation. The ways of constructing and employing the face prior can be further explored for a more realistic generation.

\bibliographystyle{siamplain}
\bibliography{refs}

\end{document}